\documentclass{article}

\usepackage{arxiv}

\usepackage[utf8]{inputenc} 
\usepackage[T1]{fontenc}    
\usepackage{hyperref}       
\usepackage{url}            
\usepackage{booktabs}       
\usepackage{amsfonts}       
\usepackage{nicefrac}       
\usepackage{microtype}      
\usepackage{lipsum}		
\usepackage{graphicx}
\usepackage{doi}

\usepackage{comment}
\usepackage{bm}
\usepackage{graphicx}
\usepackage{acronyms}
\usepackage{multirow}
\usepackage[export]{adjustbox}
\usepackage[labelsep=quad,indention=10pt]{subfig}
\usepackage[dvipsnames]{xcolor}
\usepackage{verbatim}
\usepackage{lineno,hyperref}
\usepackage{adjustbox}
\usepackage{rotating}
\usepackage{amsmath}
\usepackage{amsfonts}
\usepackage{algorithmic}
\usepackage{algorithm}
\usepackage{amsmath}
\usepackage{amsfonts}

\title{Data-driven controllers and the need for perception systems in underwater manipulation}

\date{} 					

\author{ 
    {James P. Oubre} \\
	\textit{Department of Mechanical \& Industrial Engineering} \\
	\textit{Louisiana State University} \\
     Baton Rouge, USA  \\
	\texttt{joubr13@lsu.edu} \\
	\And
    {Ignacio Carlucho} \\
	\textit{Department of Mechanical \& Industrial Engineering} \\
	\textit{Louisiana State University} \\
     Baton Rouge, USA  \\
	\texttt{icarlucho@lsu.edu} \\
	\And
	{Corina Barbalata} \\
	\textit{Department of Mechanical \& Industrial Engineering} \\
	\textit{Louisiana State University} \\
     Baton Rouge, USA  \\
	\texttt{cbarbalata@lsu.edu} \\
}

\renewcommand{\headeright}{}
\renewcommand{\undertitle}{}
\renewcommand{\shorttitle}{Data-driven Controllers and the Need for Perception Systems in Underwater Manipulation}

\hypersetup{
pdftitle={Data-driven controllers and the need for perception systems in underwater manipulation},
pdfsubject={Robotics},
pdfauthor={James P. Oubre, Ignacio Carlucho, Corina Barbalata},
pdfkeywords={Underwater manipulation, Perception, Autonomous underwater vehicle},
}

\begin{document}
\maketitle

\begin{abstract}
The underwater environment poses a complex problem for developing autonomous capabilities for \acp{UVMS}. The modeling of \acp{UVMS} is a complicated and costly process due to the highly nonlinear dynamics and the presence of unknown hydrodynamical effects.
This is aggravated in tasks where the manipulation of objects is necessary, as this may not only introduce external disturbances that can lead to a fast degradation of the control system performance, but also requires the coordinating with a vision system for the correct grasping and operation of the object.  
In this article, we introduce a control strategy for \acp{UVMS} working with unknown payloads. The proposed control strategy is based on a data-driven optimal controller. 
We present a number of experimental results showing the benefits of the proposed strategy. Furthermore, we include a discussion regarding the visual perception requirements for the \ac{UVMS} in order to achieve full autonomy in underwater manipulation tasks of unknown payloads. 

\end{abstract}

\section{INTRODUCTION}
Currently, there are increasing efforts to explore and exploit maritime resources to further expand and advance capabilities within the oil \& gas industry, research, and military applications.
However, many of these underwater applications consist of intervention tasks in which dexterous manipulation is required \cite{SANZ2010187}. Usually, a \ac{ROV} will perform these tasks where an underwater manipulator is attached to a vehicle, and a human operator is in charge of driving the system. This approach involves high operational costs due to the need of specialized operators and large infrastructures.  An alternative solution is the use of fully autonomous systems \cite{MATSUDA2019103231,JonatanAUV}, where a manipulator is mounted on an \ac{AUV}. With these systems, underwater manipulators can perform autonomous tasks in any underwater environment \cite{HAUGALOKKEN20181}. A key stepping-stone in developing fully autonomous capabilities for manipulators on board \acp{AUV} is implementing robust control structures and active perception systems. This is challenging as these systems are high-dimensional, highly nonlinear, governed by parametric uncertainties, have limited power supply, and they have to act robustly in environments with very poor visibility \cite{Barbalata2014Coupling}.

\begin{figure}[t]
  \includegraphics[width=\linewidth]{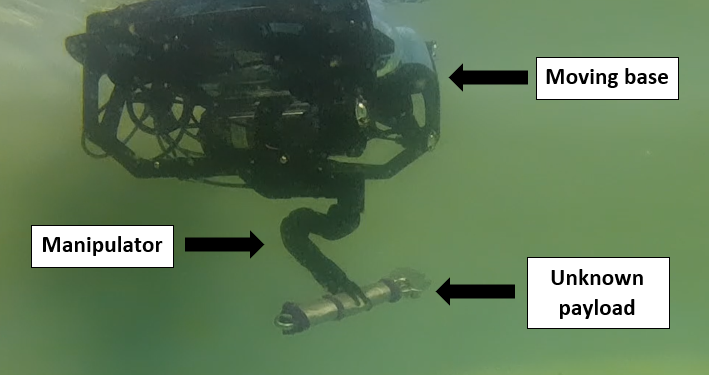}
  \caption{ The underwater manipulator Reach Alpha 5 is mounted on the BlueROV2 Heavy and has a wrench as payload. }
  \label{fig:underwater_system}
\end{figure}

Underwater intervention tasks usually require interaction with the environment and manipulation of payloads that might be unknown a priori. For instance, some tasks may require the underwater manipulator system to carry or transfer some payload, or to use specialized tools for maintenance/repair applications  \cite{SIVCEV2018431}. This increases the difficulty of the manipulation problem as the system has to correctly localize and grasp these objects. Furthermore, once these objects have been grasped the behavior of the manipulator system is now affected by the dynamics of the objects. 
In addition, information about the possible payloads may be unknown which further increases the problem's complexity \cite{SONG199159}. 
In search of complete autonomy for the manipulation of unknown payloads in underwater environments, there is an express need of having a reliable perception system but at the same time a 
control system robust to changes and capable of performing complex operations is required. However, this problem has not been extensively studied in the literature \cite{AdaptivePID-payload}. 
While the paper starts by presenting the control architecture for \ac{UVMS}, Fig. \ref{fig:underwater_system},  working with such unknown payloads, it also provides a detailed discussion about the perception system needed to achieve high accuracy underwater manipulation autonomously.

\section{Underwater manipulator dynamics}
\label{sec:problem}

The dynamic model of an underwater manipulator is given by the equation of motion, as follows: 
\begin{equation}
  M(\mathbf{q}) \ddot{\mathbf{q}} + C(\mathbf{q}, \dot{\mathbf{q}} ) \dot{\mathbf{q}} +  D(\mathbf{q}, \dot{\mathbf{q}} ) \dot{\mathbf{q}} + \eta(\mathbf{q}) + f_f(\mathbf{q})= \boldsymbol{\tau } - J^T F
  \label{eq:dyn_model}
\end{equation}

\noindent  where $\boldsymbol{\tau } \in \mathbb{R}^{n}$ are the torques of the manipulator joints, and $n$ is the number of joints of the manipulator, $\mathbf{q} \in \mathbb{R}^{n}$ are the joint positions, $\dot{\mathbf{q}} \in \mathbb{R}^{n}$ represent the joint velocities, and  $\ddot{\mathbf{q}} \in \mathbb{R}^{n}$ are the joint accelerations, with $n$ being the number of joints. On the left side of the equation, we have the inertial matrix $ M(\mathbf{q}) \in \mathbb{R}^{n \times n} $, the Coriolis and centripetal vector  $C(\mathbf{q}, \dot{\mathbf{q}} )  \in \mathbb{R}^{n}$, the damping and lift vector $D(\mathbf{q}, \dot{\mathbf{q}} ) \in \mathbb{R}^{n}$, the restoring forces vector $\eta(\mathbf{q}) \in \mathbb{R}^{n}$, and the friction vector $f_f(\mathbf{q}) \in \mathbb{R}^{n}$ \cite{Fossen1994GuidanceAC,Antonelli2013}.  On the right side, 
 the vector of interaction forces with the environment is represented as  $F \in \mathbb{R}^{6}$  with $J \in \mathbb{R}^{6}$ being the Jacobian of the manipulator \cite{Featherstone}.

While some parameters in the model presented in Eq. \eqref{eq:dyn_model} can be easily obtained, others, such as those related to hydrodynamic effects, can be difficult to obtain \cite{Barbalata2018CoupledAD}.
Consequently, simplifications are usually made about certain parameters while obtaining the model. Furthermore, learning the information of the environmental forces requires force/torque sensors, which can be costly, especially for deep-sea applications. While it is possible to estimate the interaction force vector using current measurements from the manipulator, they tend to have a high degree of uncertainty \cite{BARBALATA2018150}.

An alternative to classical modeling comes from data-driven techniques.
In these formulations, a black box type function is obtained that relates the input forces of the system, $\boldsymbol{\tau }$, with the resulting position and velocities \cite{Lambert_NNMODEL}. Under such formulations, artificial neural networks or Gaussian processes are used to learn the dynamics of the system \cite{Haykin1998}. The advantage of this type of modeling is that the model learns the true interaction without any simplifications to the model. Under such models, the dynamics of the system can be written as: 
\begin{equation}
[\mathbf{q}_{t+1}, \dot{\mathbf{q}}_{t+1}]  = \boldsymbol{g}( \mathbf{q}_t, \dot{\mathbf{q}}_t, \boldsymbol{\tau }_t | \, \boldsymbol{\theta})
\label{eq:data_driven_model}
\end{equation}  
\noindent where $\boldsymbol{g}$ is the \ac{NN} function and $\boldsymbol{\theta}$ are the parameters of the network. Through this technique, obtaining the robotic system's model  becomes the task of finding the \ac{NN} parameters that best describe the relationship of the input data to the output data \cite{SMARRA20181252}.

While data-driven techniques may provide an improvement over classical modeling, some problems still persist. Changes between the training and testing configuration of the manipulator is one of the limitations of data-driven models. For example, if in the testing phase the robot will have to manipulate unknown payloads, this can cause discrepancies between the \ac{NN} model obtained during training phase and the one used in the testing phase. 
These changes need to be taken into account within the model or a loss of accuracy occurs, which may cause a performance degradation. However, in order to do so, it would be necessary to have information about the objects a priori, which is not always possible.
Furthermore, when working with small manipulators, such as the one in this article, the weight of payloads may be comparable to the actual size of the manipulator.
For example, the Reach 5 Alpha manipulator weights only 1.3 kilograms but can lift up to 2 kilograms underwater. 

Therefore, when working with unknown payloads, data-driven techniques are not enough to ensure the desired working conditions will be met. In the following sections, we will develop an adaptive formulation within the context of optimal control that together with the data-driven modeling approach shows promising results for the control of underwater manipulators.

\section{Control System for Underwater Manipulators}

\begin{figure*}[!ht]
\centering
  \includegraphics[width=0.9\textwidth]{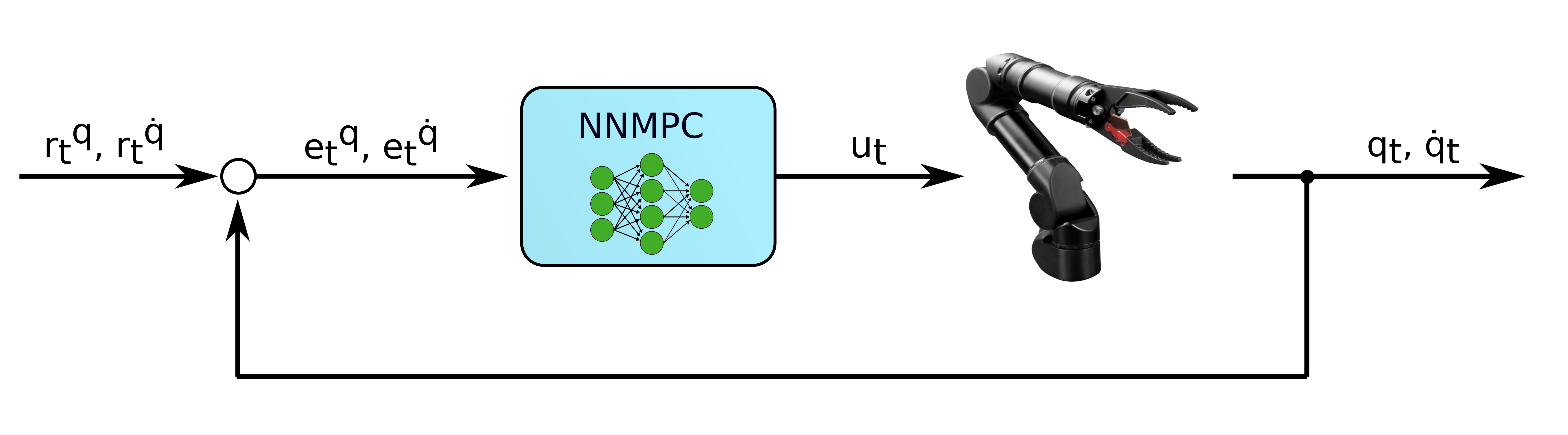}
  \caption{ Basic structure of the proposed \ac{NNMPC} controller. The \ac{NNMPC} takes as input the error between the current state of the system ($\dot{\mathbf{q}}_t$ and $\mathbf{q}_t$) and the desired reference ($\mathbf{r}_t^{\mathbf{q}}$ and $\mathbf{r}_t^{\dot{\mathbf{q}}}$) and outputs the control commands to the arm ($\mathbf{u}_t$). }
  \label{fig:basic_structure}
\end{figure*}

In this section, we will describe the main components of our proposed algorithm, the \ac{NNMPC} controller. 
The basic control system structure can be seen in Fig.  \ref{fig:basic_structure}. Our control system is based on a two layer architecture. In the lower layer (Data-Driven Model Predictive Control), we have a data-driven \ac{MPC} controller, namely the \ac{NNMPC}. This controller uses a data-driven model based on a neural network. The controller receives the current state of the system ($\dot{\mathbf{q}}_t$ and $\mathbf{q}_t$) and a desired reference ($\mathbf{r}_t^{\mathbf{q}}$ and $\mathbf{r}_t^{\dot{\mathbf{q}}}$), and outputs the optimal action ($\mathbf{u}_t$) that will bring the arm to the desired reference state. In the following paragraphs, we introduce the lower layer of the controller.

\subsection{Data-Driven Model Predictive Control}

The \ac{MPC} control approach focuses on minimizing a cost function over a finite horizon window. This minimization takes into consideration any constraints imposed on the system, such as the joint limits of the manipulator and the dynamic model of the system. 
In this project, we define a cost function in the context of \ac{MPC}, as:

\begin{equation} \label{eq:costf}
\begin{split}
J(t) = & {P(t)} \sum_{t_0}^{t} \mathbf{e}_t \Delta l + \sum_t^{t+N} ||\mathbf{r}_t^{\mathbf{q}} - \mathbf{q}_t||^2_{Q_1(t)} \\ + &  ||\mathbf{r}_t^{\dot{\mathbf{q}}} -\dot{\mathbf{q}}_t||^2_{Q_2(t)} + ||\Delta \mathbf{u}_t||^2_{R(t)}
\end{split}
\end{equation}

\noindent where $\mathbf{e}_t \in \mathbb{R}^{n}$ is the integral error in position with $ \Delta l \in \mathbb{R}$ being the integration step,  

$\mathbf{r}_t^{\mathbf{q}} \in \mathbb{R}^{n}$ and $\mathbf{r}_t^{\dot{\mathbf{q}}}  \in \mathbb{R}^{n}$ are the position and velocity references, $\mathbf{q}_t \in \mathbb{R}^{n}$ and $\dot{\mathbf{q}_t}  \in \mathbb{R}^{n}$ are the position and velocity states, $\mathbf{u}_t \in \mathbb{R}^{n}$ is the torque commands sent to the arm, and  $\Delta \mathbf{u}_t \in \mathbb{R}^{n}$ is the variation between commands sent at time $t$ and $t-1$. Additionally, $ R(t) \in \mathbb{R}^{n\times n}$, $Q_1(t) \in \mathbb{R}^{n\times n}$, $Q_2(t) \in \mathbb{R}^{n\times n}$ and $P(t) \in \mathbb{R}^{n\times n}$ are the weight matrices, and $N$ is the horizon window for which the problem is being solved. Notice that we make all the weight matrices ($R(t)$, $Q_1(t)$, $Q_2(t)$, $P(t)$) dependent on time, which means that these matrices can be modified depending on the current state of the system at time $t$.

In the classical \ac{MPC} formulation, the future states of the system, $ \mathbf{s}_{t+1} = [\mathbf{q}_{t+1}, \dot{\mathbf{q}}_{t+1}]$ to \( \mathbf{s}_{t+N} = [\mathbf{q}_{t+N}, \dot{\mathbf{q}}_{t+N}]\), are typically predicted using a linear model of the plant being controlled. In our case, however, we are interested in developing a data-driven formulation using neural networks. Under this scenario, we consider a neural network that will predict the successive states of the system as: 
\begin{equation} \label{eq:dd_model}
[\Delta  \mathbf{q}_{t+1}, \Delta  \dot{\mathbf{q}}_{t+1}] = \boldsymbol{g}( \mathbf{q}_{t},  \dot{\mathbf{q}}_{t} , \mathbf{u}_t \, | \, \boldsymbol{\theta} )
\end{equation}   
\noindent where $\boldsymbol{g}(\cdot) \in \mathbb{R}^{3n} \rightarrow \mathbb{R}^{2n} $ is the neural network function, $ \boldsymbol{\theta}$ are the parameters of the \ac{NN}, $\Delta\mathbf{q}_{t+1} \in \mathbb{R}^{n}$ is the change in position, and $\Delta \dot{\mathbf{q}}_{t+1} \in \mathbb{R}^{n}$ is change in velocity. With this formulation, instead of directly learning the successive state of the system, we learn the change with respect to the previous state \cite{Lambert_NNMODEL}. The successive state can then be obtained by a simple operation, i.e., $\mathbf{q}_{t+1} = \mathbf{q}_t + \Delta \mathbf{q}_{t+1}$.

With this formulation in mind, the \ac{MPC} control formulation can be written in the following way: 
\begin{equation} \label{eq:MPC}
\begin{split}
\mathbf{u^*(t)} = \min_{\boldsymbol{\mathbf{u}}_{t+ \ell}}  & {P(t)} \sum_{t_0}^{t} \mathbf{e}_t \Delta l + \sum_t^{t+N} ||\mathbf{r}_t^{\mathbf{q}} - \mathbf{q}_t||^2_{Q_1(t)} \\ + & ||\mathbf{r}_t^{\dot{\mathbf{q}}} -\dot{\mathbf{q}}_t||^2_{Q_2(t)} + ||\Delta \mathbf{u}_t||^2_{R(t)}
\end{split}
\end{equation} 

\vspace{-12pt}
\begin{flalign*}
\quad \quad \text{   subject to } \quad & \mathbf{u}_{t+ \ell} \in \mathbb{U}^n & \\
  & \textbf{q}_{t+ \ell} \in \mathbb{X}^{n}  & \\
  & \forall \, t=0, ..., N &
\end{flalign*}

\noindent where the optimal action ($\mathbf{u^*(t)}$) is obtained by minimizing the cost function $J(t)$ in Eq. \eqref{eq:costf}, and the position and velocity at time $t$ are obtained using the \ac{NN} formulation previously described in Eq. \eqref{eq:dd_model}.

Although the data-driven formulation provides advantages in comparison to the classical \ac{MPC}, large variations in the dynamics of the system can create problems where the data-driven approach may not be able to ensure the control requirements are met. Under such cases, it may be necessary to re-tune the parameters of the control system to adjust to the changing dynamics, as one set of tuning parameters may not be enough for every working condition.
This brings the need for an adaptation mechanism that allows the weights of the \ac{MPC} to be modified on-line.

\begin{figure*}[!ht]
        \centering
           \subfloat[NNMPC holding Wrench 2]{%
              \includegraphics[trim=20 5 20 5, width=0.5\textwidth]{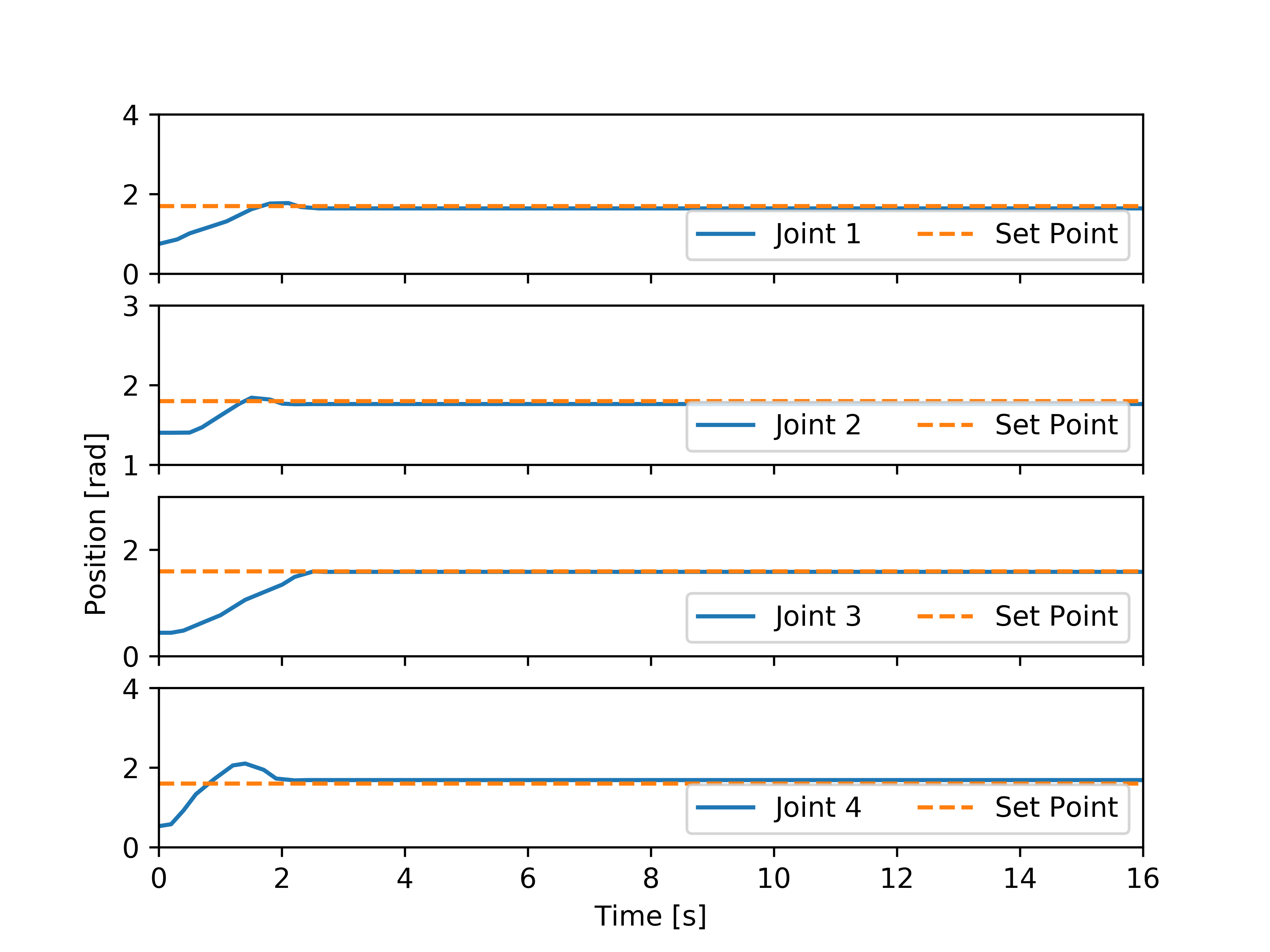}%
              \label{fig:adannmpc_wrench2}%
           } \hfil
            \subfloat[NNMPC holding Weights]{%
              \includegraphics[trim=20 5 20 5, width=0.5\textwidth]{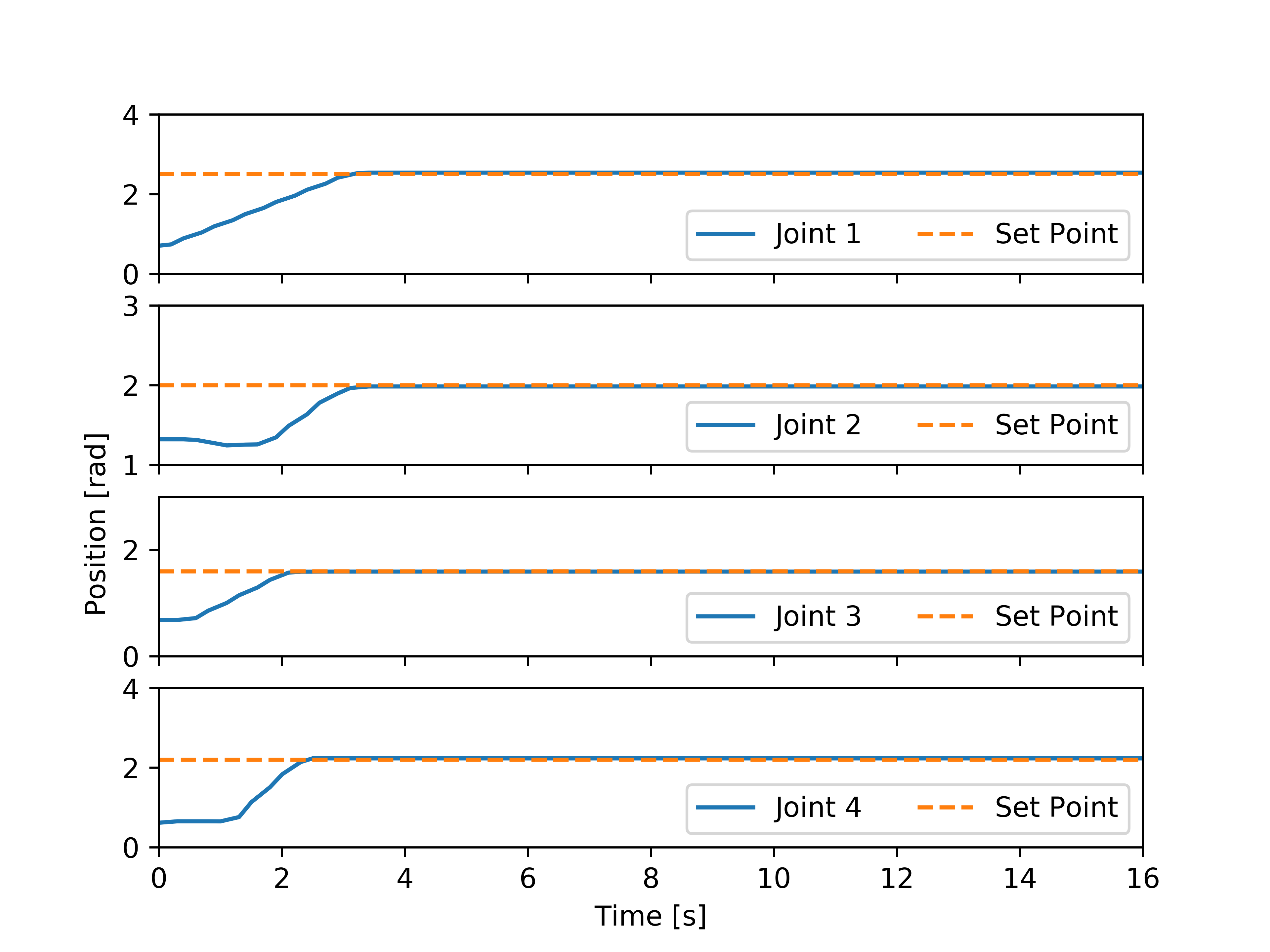}%
              \label{fig:adannmpc_weights}%
           } \hfil
           \caption{NNMPC controller on moving base holding different weights. a) Holding the  wrench with ${\mathbf{r}_t = [1.7, 1.8, 1.6, 1.6]}$ radians. b) Holding weights with $\mathbf{r}_t = [1.7, 1.8, 1.6, 1.6]$ radians }
           \label{fig:adannmpc_moving_base}
\end{figure*}

\section{Controller evaluation}

The controller presented in the previous section has been deployed on the Reach Alpha 5 underwater manipulator  \cite{reachalpha} capable of working in depths of up to 300m. This manipulator has five \acp{DOF} consisting of four rotating joints and one end-effector gripper. However, we are not interested in controlling the gripper so for the remainder of the results section we will be working with the first four \ac{DOF}. 
We consider the case of the arm moving once the unknown payload was already grabbed by the arm, so a vision system for detecting the object was not incorporated. 
We utilize \ac{ROS} to communicate with the manipulator at a frequency of 20Hz.

The data-driven \ac{NN} model of the manipulator was developed using Tensorflow. The employed artificial neural network was a fully connected feed forward net with two hidden layers of 25 neurons each. The input to the neural network is a vector of size 12 consisting of the positions ($\mathbf{q}_t$), velocities ($\mathbf{q}_t$) and actions ($\mathbf{u}_t$). The net provides the estimated successive state at time $t+1$. The data for training the network was obtained with real-time data, by performing random motions with the manipulator. The activation functions used for the hidden layers are \textit{ReLU} \cite{Vinod_relu}, while a \textit{tanh} function is used in the output layer. In the implementation of the \ac{NN}, we utilized the Mean Squared Error Loss (MSE) function, which has an equation $L(\boldsymbol{\theta}) = \sum_i^N (\mathbf{s}_i - \hat{y}_i)^2$. The training of the neural network was performed using the Adam optimizer, a gradient based optimization algorithm \cite{Kingma2015AdamAM} available in the Tensorflow library. Additionally, we used a learning rate of $lr = 0.001$.

The cost function used for all experiments is that of Eq. \eqref{eq:costf}. 
To solve the \ac{MPC} optimization problem, we used the Python library pyOpt \cite{pyopt-paper}, and we utilized the \ac{COBYLA} algorithm \cite{Powell1994}. We set an horizon window of $N=7$ with a sampling rate of $dt=0.05$ seconds.

 The tests were performed in a cylindrical water tank with a depth of 1.5 meters and a diameter of 2.5 meters.
In these tests, we are interested in the performance of the algorithm when the arm is manipulating a variety of objects. 
We show results for two different objects, each with different characteristics, that the arm has to hold with the gripper while trying to reach the desired joint positions. 
Information about the weight and geometry of the objects was unavailable to the control systems at all times, and therefore these objects are effectively unknown. Therefore, for the remainder of the paper we refer to these objects as unknown payloads.
The payloads are: 
a wrench weighing 0.5 kg (wrench) and 12 inches long,  %
and a set of weights packed together weighing 1 kg (weights) and three inches width and length and two inches height.
However, when selecting the objects, the dry mass range of the payloads were restricted by the limitations of the Reach Alpha 5 manipulator, which has a lift capacity of 2kg \cite{reachalpha}. The testing setup can be observed in Fig. \ref{fig:underwater_system}. 

 In the first test, the arm was manipulating the wrench while trying to reach a desired joint position of $\mathbf{r}_t = [1.7, 1.8, 1.6, 1.6]$ radians. The obtained joint positions are shown in Fig. \ref{fig:adannmpc_wrench2}. It can be seen, that while the joint positions are reached in less than 3 seconds,  joint 4 presents a small overshoot, which relates to the extra torque caused by wrench in the joint. Following, we performed a test with the arm holding the 1 kg weight. For this test, the desired joint position is $\mathbf{r}_t = [2.5, 2.0, 1.6, 2.2]$ radians, and the obtained results can be seen in Fig. \ref{fig:adannmpc_weights}. We can see that due to the extreme characteristic of the payload, joints 2 and 4 require a longer time to reach the desired joint position.

\section{Perception for underwater manipulation}

The preceding results show the advantages of data-driven modeling techniques. In this regard, the \ac{NN} helps to capture unmodeled dynamics in the manipulator, which improves the overall performance of the controller.
This process brings substantial performance improvements when facing unknown payloads. This is of great importance in furthering autonomous operations that require manipulation with unknown payloads.

Regarding the sensing techniques utilized, the Reach Alpha 5 is a high precision arm that utilizes encoders for obtaining the joint states. The arm uses magnetic off-axis position encoders (iC-MU from Haus integrated circuits) which have an accuracy of $\pm 0.1$ degrees in the joint position. Those sensors are integrated inside the arm so the controller gets an immediate update with no interference.
For our current implementation, the biggest concern with regard of sensing comes from the potential case in which one of the encoders fails. In such case, the uncertainty in the measurements goes outside of the bounds given by the manufacturer and the controller will be unable to perform its required tasks.

Additionally, we only consider the case of the arm moving once the unknown payload was already grabbed by the arm, so a vision system for detecting the object was not implemented. However, this will be required for the whole system in the case of autonomous control. 
We theorize that a combination of 3D vision, utilizing cameras, together with underwater image restoration, would allow the \ac{UVMS} to obtain the necessary visual feedback to close the perception-action cycle \cite{CognitiveControlHaykin}. 

In this sense, stereo vision is a common way of controlling a robotic manipulator with camera feedback \cite{visionManipulationReview}. Typically, object detection is used to find the object that must be manipulated. Once the object is found, the next step is to find the distance to the object. This can be calculated using trigonometry by having two cameras that are a known distance apart from each other and both looking at the object \cite{3DvisionLuu}. This distance can then be sent to the control system to modify the required end-effector position based on the current payload location with respect to the \ac{UVMS} \cite{IZAGIRRE2021102029}.

However, one problem faced in underwater environments is the poor visibility, caused by the strong absorption and scattering effects \cite{DAI2020105947}. Additionally, light changes and water turbidness can cause adverse effects on the image, distorting it and decreasing the clarity. One solution to this problem could be haze removal and color correction. Haze removal makes the output image clearer and rids much of the blurriness that occurs. Next, color correction is done to extract the true colors of the image. In underwater environments, the blue and green channels present higher intensities than the red channel \cite{ParameterCompLiu}. Therefore, by performing color correction, the colors appear more vividly, as they would out of water, adding more detail to the image \cite{RestorationGUO}. 

By combining these methods, we believe that the problem of unknown payload manipulation in \ac{UVMS} could be solved autonomously. Two cameras could be used to give feedback to the controller and operate the arm correctly. Simultaneously, haze removal and color correction will modify the images from the camera to produce a more accurate representation of the underwater environment. However, in order to deploy this type of system completely autonomous, increased computational power is necessary. For this reason, we plan to increase the load capabilities of the \ac{UVMS} by incorporating another enclosure, that will allow us to mount additional cameras, together with an additional computer. We plan to incorporate an NVIDIA Jetson nano, due to its small footprint and high computational power.

\section{CONCLUSIONS}

In this work, we developed a low-level controller based on data-driven Model Predictive Control for the control of an underwater manipulator working with unknown payloads. We utilized a \ac{NN} to derive a model for the manipulator which was used by the \ac{MPC} controller. In this way, we were able to obtain a more accurate model of the manipulator that directly takes into consideration the environmental disturbances. While previous works have utilized such a formulation, most have been done in simulation or utilized the network to directly learn the commands of the \ac{MPC} controller. We were able to solve the data-driven optimization problem online by utilizing a low impact \ac{NN}.
Moreover, we provide a discussion regarding the necessary steps for the creation of a perception system that will allow the \ac{UVMS} to perform the unknown payload manipulation, in changing underwater environments, autonomously.





\section*{ACKNOWLEDGMENT}
This work was supported by Defense Advanced Research Projects Agency (DARPA) Angler program award number SC-19-027~$/$~10434.01 (AWD-48999-1). The authors would like to thank Dr. Lilia Moshkina and Dr. Eric Martinson for their comments and feedback in the development of this work.

\bibliographystyle{unsrt}
\bibliography{references}

\begin{thebibliography}{10}

\bibitem{SANZ2010187}
Pedro~J. Sanz, Pere Ridao, Gabriel Oliver, Claudio Melchiorri, Giuseppe
  Casalino, Carlos Silvestre, Yvan Petillot, and Alessio Turetta.
\newblock Trident: A framework for autonomous underwater intervention missions
  with dexterous manipulation capabilities.
\newblock {\em IFAC Proceedings Volumes}, 43(16):187 -- 192, 2010.
\newblock 7th IFAC Symposium on Intelligent Autonomous Vehicles.

\bibitem{MATSUDA2019103231}
Takumi Matsuda, Toshihiro Maki, Kotohiro Masuda, and Takashi Sakamaki.
\newblock Resident autonomous underwater vehicle: Underwater system for
  prolonged and continuous monitoring based at a seafloor station.
\newblock {\em Robotics and Autonomous Systems}, 120:103231, 2019.

\bibitem{JonatanAUV}
J.~S. {Willners}, L.~{Toohey}, and Y.~{Petillot}.
\newblock Sampling-based path planning for cooperative autonomous maritime
  vehicles to reduce uncertainty in range-only localization.
\newblock {\em IEEE Robotics and Automation Letters}, 4(4):3987--3994, 2019.

\bibitem{HAUGALOKKEN20181}
Bent Oddvar~A. Haugaløkken, Erlend~K. Jørgensen, and Ingrid Schjølberg.
\newblock Experimental validation of end-effector stabilization for underwater
  vehicle-manipulator systems in subsea operations.
\newblock {\em Robotics and Autonomous Systems}, 109:1 -- 12, 2018.

\bibitem{Barbalata2014Coupling}
C.~{Barbalata}, M.~W. {Dunnigan}, and Y.~{Pétillot}.
\newblock Dynamic coupling and control issues for a lightweight underwater
  vehicle manipulator system.
\newblock In {\em 2014 Oceans - St. John's}, pages 1--6, 2014.

\bibitem{SIVCEV2018431}
Satja Sivčev, Joseph Coleman, Edin Omerdić, Gerard Dooly, and Daniel Toal.
\newblock Underwater manipulators: A review.
\newblock {\em Ocean Engineering}, 163:431 -- 450, 2018.

\bibitem{SONG199159}
Y.D. Song and J.N. Anderson.
\newblock Adaptive path tracking control of robotic manipulators with unknown
  payload dynamics.
\newblock {\em Systems {\&} Control Letters}, 17(1):59 -- 70, 1991.

\bibitem{AdaptivePID-payload}
J.~{Lee}, P.~H. {Chang}, B.~{Yu}, and M.~{Jin}.
\newblock An adaptive pid control for robot manipulators under substantial
  payload variations.
\newblock {\em IEEE Access}, 8:162261--162270, 2020.

\bibitem{Fossen1994GuidanceAC}
T.~Fossen.
\newblock Guidance and control of ocean vehicles.
\newblock 1994.

\bibitem{Antonelli2013}
Gianluca Antonelli.
\newblock {\em Underwater Robots}.
\newblock Springer Publishing Company, Incorporated, 3rd edition, 2013.

\bibitem{Featherstone}
Roy Featherstone.
\newblock {\em Rigid Body Dynamics Algorithms}.
\newblock Springer-Verlag, Berlin, Heidelberg, 2007.

\bibitem{Barbalata2018CoupledAD}
Corina Barbalata, Matthew~W. Dunnigan, and Yvan~R. Petillot.
\newblock Coupled and decoupled force/motion controllers for an underwater
  vehicle-manipulator system.
\newblock 2018.

\bibitem{BARBALATA2018150}
Corina Barbalata, Matthew~W. Dunnigan, and Yvan Petillot.
\newblock Position/force operational space control for underwater manipulation.
\newblock {\em Robotics and Autonomous Systems}, 100:150 -- 159, 2018.

\bibitem{Lambert_NNMODEL}
N.~O. {Lambert}, D.~S. {Drew}, J.~{Yaconelli}, S.~{Levine}, R.~{Calandra}, and
  K.~S.~J. {Pister}.
\newblock Low-level control of a quadrotor with deep model-based reinforcement
  learning.
\newblock {\em IEEE Robotics and Automation Letters}, 4(4):4224--4230, 2019.

\bibitem{Haykin1998}
Simon Haykin.
\newblock {\em Neural Networks: A Comprehensive Foundation}.
\newblock Prentice Hall PTR, USA, 2nd edition, 1998.

\bibitem{SMARRA20181252}
Francesco Smarra, Achin Jain, Tullio [de Rubeis], Dario Ambrosini, Alessandro
  D’Innocenzo, and Rahul Mangharam.
\newblock Data-driven model predictive control using random forests for
  building energy optimization and climate control.
\newblock {\em Applied Energy}, 226:1252 -- 1272, 2018.

\bibitem{reachalpha}
Blueprint Lab.
\newblock Reach alpha 5 (2020).
\newblock {\em URL https://blueprintlab.com/products/reach-alpha/}, 2020.

\bibitem{Vinod_relu}
Vinod Nair and Geoffrey~E. Hinton.
\newblock Rectified linear units improve restricted boltzmann machines.
\newblock In {\em Proceedings of the 27th International Conference on
  International Conference on Machine Learning}, ICML’10, page 807–814,
  Madison, WI, USA, 2010. Omnipress.

\bibitem{Kingma2015AdamAM}
Diederik~P. Kingma and Jimmy Ba.
\newblock Adam: A method for stochastic optimization.
\newblock {\em CoRR}, abs/1412.6980, 2015.

\bibitem{pyopt-paper}
Ruben~E. Perez, Peter~W. Jansen, and Joaquim R. R.~A. Martins.
\newblock py{O}pt: A {P}ython-based object-oriented framework for nonlinear
  constrained optimization.
\newblock {\em Structures and Multidisciplinary Optimization}, 45(1):101--118,
  2012.

\bibitem{Powell1994}
M.~J.~D. Powell.
\newblock {\em A Direct Search Optimization Method That Models the Objective
  and Constraint Functions by Linear Interpolation}, pages 51--67.
\newblock Springer Netherlands, Dordrecht, 1994.

\bibitem{CognitiveControlHaykin}
Simon Haykin, Mehdi Fatemi, Peyman Setoodeh, and Yanbo Xue.
\newblock Cognitive control.
\newblock {\em Proceedings of the IEEE}, 100(12):3156--3169, 2012.

\bibitem{visionManipulationReview}
Koichi Hashimoto.
\newblock A review on vision-based control of robot manipulators.
\newblock {\em Advanced Robotics}, 17(10):969--991, 2003.

\bibitem{3DvisionLuu}
Trong~Hieu Luu and Thanh~Hung Tran.
\newblock 3d vision for mobile robot manipulator on detecting and tracking
  target.
\newblock In {\em 2015 15th International Conference on Control, Automation and
  Systems (ICCAS)}, pages 1560--1565, 2015.

\bibitem{IZAGIRRE2021102029}
Unai Izagirre, Imanol Andonegui, Luka Eciolaza, and Urko Zurutuza.
\newblock Towards manufacturing robotics accuracy degradation assessment: A
  vision-based data-driven implementation.
\newblock {\em Robotics and Computer-Integrated Manufacturing}, 67:102029,
  2021.

\bibitem{DAI2020105947}
Chenggang Dai, Mingxing Lin, Xiaojian Wu, Zhen Wang, and Zhiguang Guan.
\newblock Single underwater image restoration by decomposing curves of
  attenuating color.
\newblock {\em Optics \& Laser Technology}, 123:105947, 2020.

\bibitem{ParameterCompLiu}
Junnan Liu and Xiaoyu Zhang.
\newblock Parameter-adaptive compensation (pac) for processing underwater
  selective absorption.
\newblock {\em IEEE Signal Processing Letters}, 27:2178--2182, 2020.

\bibitem{RestorationGUO}
Jun-Kai Guo, Chia-Chi Sung, and Herng-Hua Chang.
\newblock Restoration of underwater vision using a two-phase regularization
  mechanism.
\newblock In {\em 2014 7th International Congress on Image and Signal
  Processing}, pages 243--247, 2014.

\end{thebibliography}







\end{document}